\documentclass[sigconf, screen]{acmart}
\AtBeginDocument{%
  }

\acmDOI{XXXXXXX.XXXXXXX}
\acmConference[ASE '26]
  {Proceedings of the 41st IEEE/ACM International Conference on Automated Software Engineering}
  {October 12--16, 2026}
  {Munich, Germany}

\usepackage{xspace}
\usepackage{hyperref}
\hypersetup{hidelinks}
\usepackage{enumitem}
\usepackage{tcolorbox}
\usepackage{subfig}
\usepackage{multirow}

\newcommand{\eg}{e.g.\xspace}

\begin{document}

\title{ARMOR: A Robust Self-Supervised Framework for Root Cause Analysis in Microservices under Missing Modality}

\author{Wenzhuo Qian}
\orcid{0009-0006-6059-2492}
\affiliation{%
  \institution{Zhejiang University}
  \city{Hangzhou}
  \country{China}
}
\email{qwz@zju.edu.cn}

\author{Hailiang Zhao}
\authornote{Corresponding authors.}
\orcid{0000-0003-2850-6815}
\affiliation{%
  \institution{Zhejiang University}
  \city{Hangzhou}
  \country{China}
}
\email{hliangzhao@zju.edu.cn}

\author{Ziqi Wang}
\orcid{0000-0001-7176-2369}
\affiliation{%
  \institution{Zhejiang University}
  \city{Hangzhou}
  \country{China}
}
\email{wangziqi0312@zju.edu.cn}

\author{Zhipeng Gao}
\orcid{0009-0002-0077-1916}
\affiliation{%
  \institution{Zhejiang University}
  \city{Hangzhou}
  \country{China}
}
\email{22551125@zju.edu.cn}

\author{Jiayi Chen}
\orcid{0009-0006-3383-3115}
\affiliation{%
  \institution{Zhejiang University}
  \city{Hangzhou}
  \country{China}
}
\email{jyichen@zju.edu.cn}

\author{Zhiwei Ling}
\orcid{0000-0001-9705-1987}
\affiliation{%
  \institution{Zhejiang University}
  \city{Hangzhou}
  \country{China}
}
\email{zwling@zju.edu.cn}

\author{Shuiguang Deng}
\authornotemark[1]
\orcid{0000-0001-5015-6095}
\affiliation{%
  \institution{Zhejiang University}
  \city{Hangzhou}
  \country{China}
}
\email{dengsg@zju.edu.cn}

\renewcommand{\shortauthors}{Qian et al.}

\begin{abstract}
Automated incident management is critical for microservice reliability. While recent unified frameworks leverage multimodal data for joint optimization, they unrealistically assume perfect data completeness. In practice, network fluctuations and agent failures frequently cause missing modalities. Existing approaches relying on static placeholders introduce imputation noise that masks anomalies and degrades performance. To address this, we propose ARMOR, a robust self-supervised framework designed for missing modality scenarios. ARMOR features: (i) a modality-specific asymmetric encoder that isolates distribution disparities among metrics, logs, and traces; and (ii) a missing-aware gated fusion mechanism utilizing learnable placeholders and dynamic bias compensation to prevent cross-modal interference from incomplete inputs. By employing self-supervised auto-regression with mask-guided reconstruction, ARMOR jointly optimizes anomaly detection (AD), failure triage (FT), and root cause localization (RCL). AD and RCL require no fault labels, while FT relies solely on failure-type annotations for the downstream classifier. Extensive experiments demonstrate that ARMOR achieves state-of-the-art performance under complete data conditions and maintains robust diagnostic accuracy even with severe modality loss.
\end{abstract}



\maketitle

\section{Introduction}

Microservice architectures have become the foundational infrastructure for modern cloud-native applications due to their inherent scalability and agility~\cite{dragoni2017microservices, sriraman2019softsku}. However, the massive scale and dynamic complexity of these distributed systems make performance anomalies and system failures inevitable, frequently leading to severe service disruptions and substantial financial losses~\cite{finance}. Consequently, ensuring system reliability through automated incident management has become a priority for site reliability engineers (SREs)~\cite{tao2024diagnosing}.

\begin{figure}[t]
\centering
\includegraphics[width=1\columnwidth]{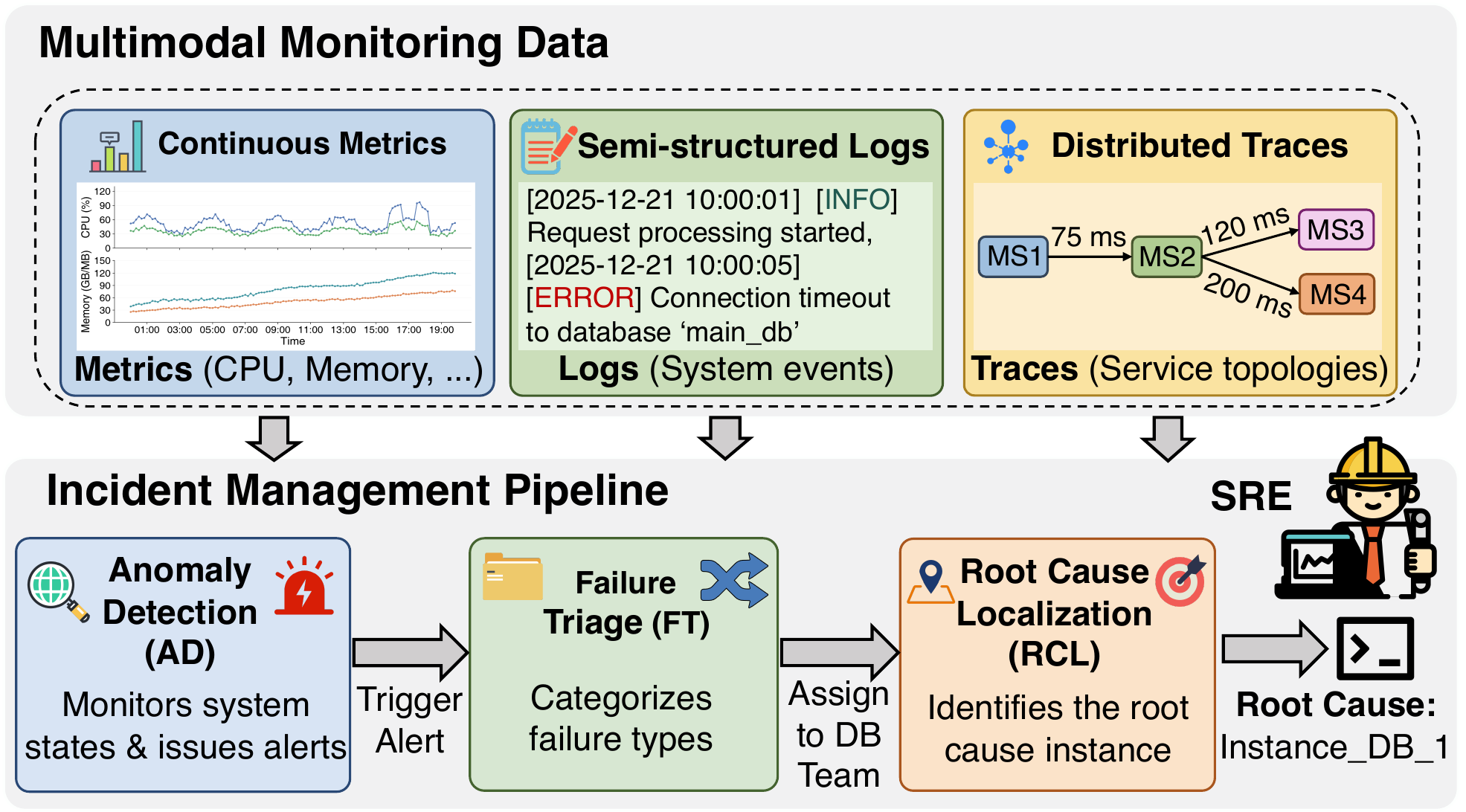}
\caption{Overview of the multimodal incident management pipeline. Diverse observability data (metrics, logs, and traces) drives three sequential tasks: AD, FT, and RCL.}
\label{fig:incident_pipeline}
\end{figure}

To manage incidents effectively, SREs rely heavily on multimodal monitoring data, including continuous metrics, semi-structured logs, and distributed traces, which collectively capture the overall system state~\cite{lee2023eadro, sun2024art, nie2025dest, sun2025trioxpert}. As Figure~\ref{fig:incident_pipeline} illustrates, these diverse data streams sequentially drive a standard incident management pipeline comprising three diagnostic tasks: anomaly detection (AD)~\cite{audibert2020usad, chen2024lara, chen2024cluster, huang2023twin, lee2023heterogeneous, zhang2025integrating, kim2025causality, cho2025structured, ding2025enhancing}, failure triage (FT)~\cite{ma2020diagnosing, liu2022microcbr, tao2024giving, sun2025failure, sui2023logkg}, and root cause localization (RCL)~\cite{somashekar2024gamma, lin2024root, pham2024root, han2025root, sun2025interpretable}. Initially, AD continuously monitors system states to trigger alerts upon detecting abnormal behaviors. Subsequently, FT categorizes the detected anomalies into specific failure types for appropriate engineering teams. Finally, RCL identifies the exact culprit instance responsible for the failure.

Current intelligent incident management approaches generally fall into two categories. Early methods rely on isolated, single-task techniques for different diagnostic stages~\cite{li2022actionable, lee2023heterogeneous, tao2024giving}. However, decoupling these related tasks ignores the shared diagnostic context embedded in system deviations, often leading to redundant maintenance overhead, inefficient resource utilization, and delayed mitigation responses~\cite{nie2025dest}. To address these limitations, recent research shifts toward unified multi-task frameworks~\cite{lee2023eadro, sun2024art, nie2025dest, sun2025trioxpert}. Instead of maintaining separate pipelines, these methods extract shared knowledge from multimodal monitoring data to jointly optimize AD, FT, and RCL. By modeling inherent dependencies across these stages, such end-to-end solutions achieve significant performance improvements and streamline troubleshooting under ideal conditions.

Despite their success, existing unified frameworks~\cite{lee2023eadro, sun2024art, nie2025dest, sun2025trioxpert} assume that the collected multimodal monitoring data is perfectly aligned and complete. However, in real-world production environments, observability infrastructure vulnerabilities, such as network fluctuations, telemetry agent crashes, and configuration errors, frequently cause missing modalities regardless of the actual microservice status~\cite{ding2025adaptive, zhang2025reconrca}. When processing incomplete data, current models typically fuse the available inputs with static placeholders~\cite{zhang2023benefit, ashok2024traceweaver}, causing significant performance degradation. Although supervised approaches attempt to handle incomplete modalities, the extreme scarcity of industrial fault labels renders them prone to overfitting and difficult to deploy. Learning the diagnostic backbone from unlabeled telemetry through self-supervision provides a practical way to reduce reliance on dense manual annotations while preserving lightweight task-specific adaptation when semantic labels are needed. Nevertheless, developing a robust, self-supervised backbone capable of supporting missing-aware incident management requires addressing three primary challenges:

\textbf{(1) Tightly coupled encoding makes any modality absence corrupt surviving signals.} When observability infrastructure fails, multiple modalities may become unavailable at the same time, and SREs cannot predict which combination will drop next. Existing methods tightly couple metrics, logs, and traces during early representation learning, so missing modalities corrupt the representations built from the survivors~\cite{lee2023eadro, sun2024art, zhang2025integrating, sun2025trioxpert}. The structural asymmetry between dense continuous metrics and sparse discrete logs and traces makes this worse: encoding heterogeneous signals with a shared architecture forces the surviving modalities into a representation space shaped by the absent ones, losing the diagnostic clues they actually carry.

\textbf{(2) Static imputation masks real failures by mimicking a healthy idle state.} When a Prometheus agent goes offline, filling its missing channels with zero tells the model that every monitored resource sits at exactly zero consumption~\cite{sun2025clusterrca, zhang2023benefit}. That is precisely the signature of a healthy idle service, not a failing one. Conventional fusion mechanisms~\cite{sun2024art, nie2025dest, zhang2025integrating} adopt rigid early concatenation and rely on static imputation~\cite{zhang2023benefit, ashok2024traceweaver} for structural consistency. The resulting pseudo-normal signal overwhelms the anomaly indicators captured by surviving logs and traces, suppressing the deviations that SREs would otherwise use to diagnose the incident.

\textbf{(3) Distorted representations break unified diagnosis across all three tasks.} Without a representation that holds up under missing data, SRE teams lose the ability to run a single unified AD, FT, and RCL pipeline. Each missing modality scenario produces a different distortion in the fused representation, forcing teams back to maintaining separate per-modality diagnostic configurations and reintroducing the fragmentation overhead that unified frameworks were designed to eliminate. Self-supervised models derive diagnostic signals from predictive errors~\cite{sun2024art, sun2025interpretable, zhang2025integrating}, but static imputation biases those errors toward normal states. Deriving a stable failure signature from unlabeled telemetry under these distorted inputs, one that remains discriminative across the full AD, FT, and RCL pipeline, remains an unresolved problem.

To address these limitations, we propose ARMOR, an Automated and Robust framework handling Missing mOdality for Root cause analysis in microservices. Specifically, ARMOR comprises three core modules designed to systematically resolve the aforementioned challenges: \textit{(1) Modality-specific status learning.} We design an asymmetric encoder tailored to the distinct distribution properties of continuous metrics and sparse events. This decoupled architecture hierarchically disentangles temporal, channel, and spatial dependencies, preventing the absence of a single modality from corrupting the representations of the others. \textit{(2) Missing-aware global fusion.} We introduce an attention-guided gating mechanism incorporating learnable placeholders and targeted biases. Instead of employing static imputation, this module explicitly recognizes absent inputs and adaptively compensates for their routing contributions, preventing static placeholders from inducing cross-modal interference. \textit{(3) Online diagnosis via unified representations.} ARMOR constructs a robust unified representation by concatenating explicit reconstruction errors with latent system embeddings. This comprehensive signature establishes a shared foundation to concurrently support AD, FT, and RCL, even under severe modality dropouts and strict label scarcity.
The main contributions of this work are as follows.
\begin{itemize}
    \item We present ARMOR, which, to the best of our knowledge, is the first self-supervised incident management framework explicitly designed to handle missing modalities in microservices, enabling label-free AD and RCL, with FT requiring only failure-type annotations for the frozen-encoder classifier.
    \item We devise a modality-specific asymmetric encoder and a missing-aware gated fusion mechanism. By leveraging learnable placeholders and dynamic bias compensation, ARMOR effectively mitigates imputation noise and prevents spurious correlations caused by incomplete inputs.
    \item Extensive experiments on benchmark microservice systems demonstrate that ARMOR achieves state-of-the-art performance under complete data settings and maintains superior diagnostic stability even under severe modality loss, significantly outperforming existing baselines.
\end{itemize}

\section{Motivation}

This section motivates the proposed framework by exploring the gap between ideal academic assumptions and complex industrial realities. Specifically, we examine the inevitability of missing modalities in production environments and demonstrate how current multimodal fusion strategies remain fundamentally vulnerable to such incomplete data.

\subsection{Why Are Missing Modalities Inevitable in Production Environments?}
\label{motivation:missing_exist}

Existing research on unified modeling relies heavily on the assumption of perfectly complete and aligned multimodal datasets (\eg, metrics, logs, and distributed traces)~\cite{sun2024art, nie2025dest, lee2023eadro}. However, such ideal conditions rarely exist in industrial cloud-native environments. Due to the scale and dynamic complexity of microservice architectures, the absence of modalities is a common occurrence rather than an infrequent edge case. This data absence is primarily driven by vulnerabilities within the observability infrastructure~\cite{zhang2025reconrca, ashok2024traceweaver, sun2025clusterrca, zhang2023benefit} (\eg, data collection and transmission pipelines) rather than by failures within the business microservices themselves. This observation is consistent with production observability practice, where telemetry is collected through best-effort pipelines. Scrape failures, sampling, buffering pressure, and ingestion limits in systems such as Prometheus~\cite{prometheus}, OpenTelemetry~\cite{opentelemetry}, and modern log pipelines~\cite{loki} can make observability data unavailable before downstream diagnosis.

\begin{figure}[t]
\centering
\includegraphics[width=1\columnwidth]{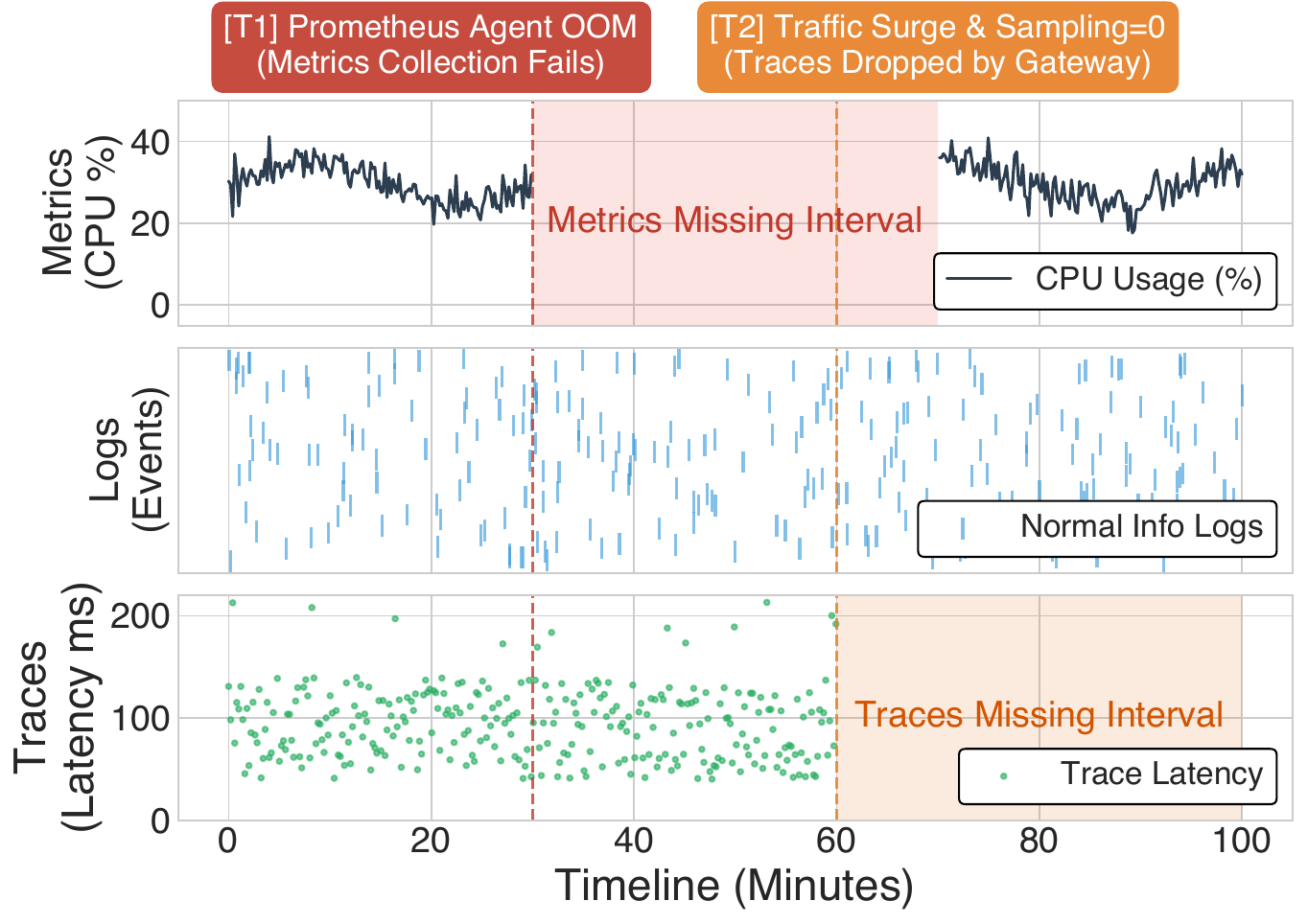}
\caption{The generation of missing modalities in production environments. This timeline illustrates how vulnerabilities in the observability infrastructure (\eg, agent crashes or aggressive sampling) cause abrupt data loss, even while the underlying microservice operates normally and outputs logs.}
\label{fig:motivation_timeline}
\end{figure}

Figure~\ref{fig:motivation_timeline} illustrates how practical factors, such as agent failures and network policies, cause abrupt data gaps even when the underlying services remain healthy. For example, telemetry agents (\eg, Prometheus \texttt{node\_exporter}) operate in resource-constrained environments and are highly susceptible to out-of-memory errors due to high monitoring overhead. At timestamp T1, an agent crash causes a sudden and prolonged absence of metrics; however, the continuous output of normal information logs confirms that the business service continues to process requests without interruption. Similarly, network fluctuations and aggressive sampling strategies create substantial data gaps. During extreme traffic surges, operations teams routinely downgrade the observability pipeline to preserve bandwidth for core transactions. At timestamp T2, this configuration results in traces being actively dropped by the gateway, yet the persistent stream of logs again verifies the operational stability of the service. Beyond such runtime interruptions, configuration errors and system heterogeneity hinder data generation. Modern systems frequently incorporate legacy services or third-party components that lack proper instrumentation, leading to an inherent absence of specific modalities without triggering explicit business errors~\cite{zhang2025reconrca, ashok2024traceweaver, sun2025clusterrca, zhang2023benefit}.

As Figure~\ref{fig:motivation_timeline} shows, missing modalities occur as complete channel-level outages that arrive irregularly across different services and time steps, primarily due to collection and transmission issues. Because extreme traffic and resource contention often precede actual system failures, the absence of observational data frequently coincides with the exact incidents that site reliability engineers must diagnose.

\begin{center}
\begin{tcolorbox}[colback=gray!10,
                  colframe=black,
                  width=8.5cm,
                  arc=5mm, auto outer arc,
                  boxrule=0.5pt]
\textbf{Finding 1:} In industrial microservices, observability infrastructure vulnerabilities produce complete modality-level blackouts: a process crash or network failure can eliminate an entire data stream instantaneously. Since such outages frequently coincide with actual system failures, methods requiring perfectly complete data are highly impractical.
\end{tcolorbox}
\end{center}

\subsection{How Do Existing Frameworks Perform with Missing Modalities?}
\label{motivation:performance}

\begin{figure}[t]
\centering
\includegraphics[width=1\columnwidth]{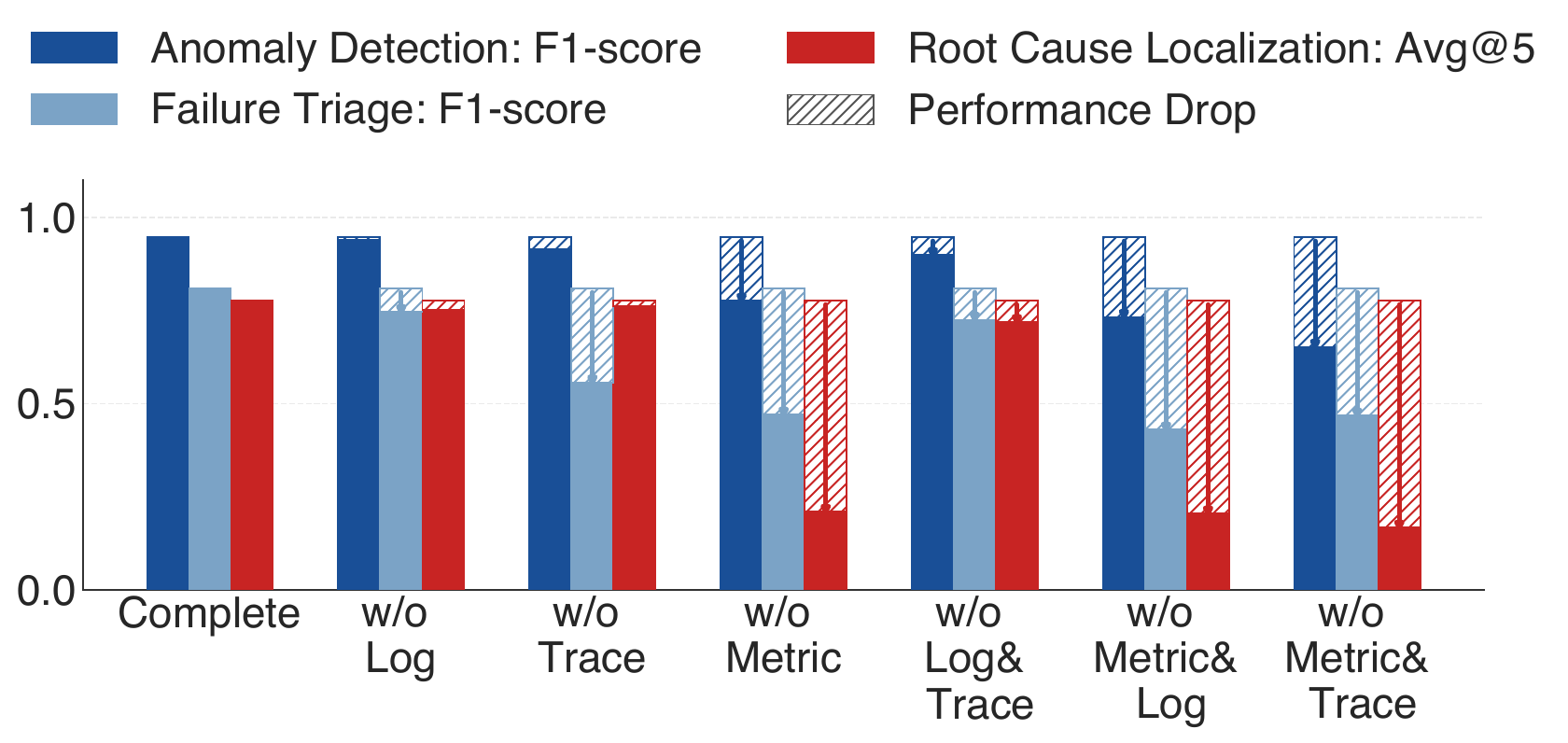}
\caption{Performance degradation of ART~\cite{sun2024art}, the strongest self-supervised unified baseline, under missing modalities. Metric loss causes a steep drop in RCL accuracy, confirming that static imputation cannot compensate for absent continuous streams.}
\label{fig:performance_degradation}
\end{figure}

Given the inevitability of missing modalities in distributed systems, evaluating how state-of-the-art unified frameworks~\cite{sun2024art, sun2025trioxpert, lee2023eadro} handle incomplete inputs is critical for practical deployment. We use ART~\cite{sun2024art}, a recent self-supervised unified incident management framework that jointly supports AD, FT, and RCL under complete observability assumptions, as the representative baseline. As Figure~\ref{fig:performance_degradation} illustrates, it experiences severe and asymmetric performance degradation under missing modalities. Although the framework maintains moderate resilience when discrete logs are missing, the absence of continuous metrics causes a significant decline, particularly in the accuracy of RCL. This result confirms that conventional static imputation fails to bridge the semantic gap left by missing modalities, leaving the automated monitoring pipeline highly vulnerable. The degradation is most severe when metrics are absent: unlike logs and traces, which record discrete events at coarse granularity, metrics provide dense, continuous resource measurements across every time step, so their loss leaves the surviving modalities with far less signal to compensate.

This systemic vulnerability stems from the rigid nature of early-concatenation fusion mechanisms. When a specific modality becomes unavailable, existing methods routinely employ static imputation (\eg, filling missing dimensions with default constants) to maintain structural consistency~\cite{zhang2025reconrca, ashok2024traceweaver, ding2025adaptive}. Although computationally convenient, this naive alignment entirely ignores semantic validity. Consequently, high-dimensional neural networks mistakenly interpret these deterministic placeholders as high-confidence features indicating a stable operational state.

\begin{figure}[t]
\centering
\includegraphics[width=0.85\columnwidth]{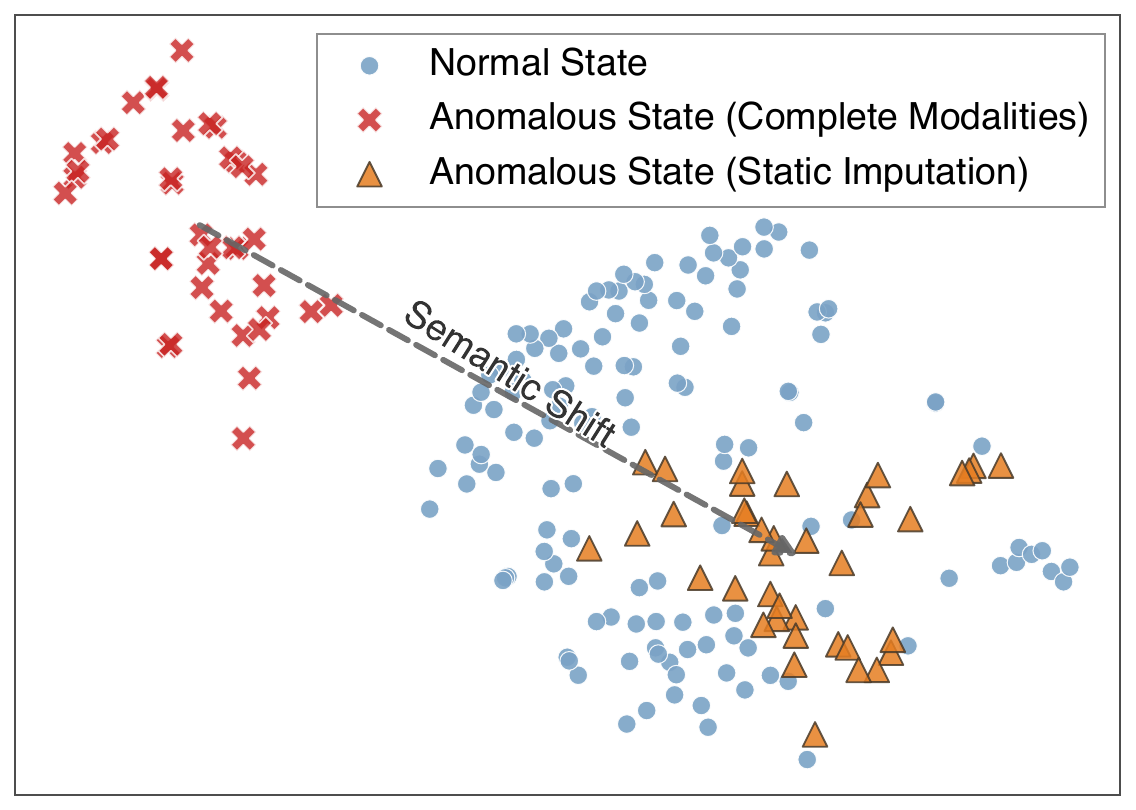}
\caption{The impact of static imputation on latent feature distributions. A t-SNE visualization demonstrates how non-adaptive default values force the features of anomalous instances into the normal cluster, creating spurious correlations that suppress actual diagnostic signals.}
\label{fig:tsne_zero_drag}
\end{figure}

This pervasive ``pseudo-normal'' signal introduces severe cross-modal interference. As the t-SNE projection in Figure~\ref{fig:tsne_zero_drag} demonstrates, static imputation forcefully shifts the latent features of anomalous instances, which otherwise form distinct and separable clusters, directly into the dense cluster of normal states. This mathematical distortion heavily suppresses the subtle anomalous deviations successfully captured by the remaining available modalities, thereby systematically reducing overall diagnostic accuracy. Therefore, eliminating this imputation-induced distortion is a prerequisite for robust incident management.

\begin{center}
\begin{tcolorbox}[colback=gray!10,
                  colframe=black,
                  width=8.5cm,
                  arc=5mm, auto outer arc,
                  boxrule=0.5pt]
\textbf{Finding 2:} The naive imputation of missing modalities biases joint representations toward normal states, severely degrading diagnostic performance and necessitating the development of robust, missing-aware fusion mechanisms.
\end{tcolorbox}
\end{center}

\section{Preliminaries}

\subsection{Self-supervised Learning}
Self-supervised learning derives supervisory signals directly from unlabeled data via pretext tasks, such as reconstruction~\cite{rumelhart1986learning, pathak2016context}, to learn generalizable representations. In microservice incident management, acquiring manual fault annotations across large-scale environments is prohibitively expensive and labor-intensive~\cite{sun2024art}. Self-supervised learning mitigates this challenge by eliminating reliance on extensive labeled data. Unlike traditional supervised joint learning requiring complex loss balancing~\cite{lee2023eadro, ding2025adaptive}, self-supervised learning naturally facilitates unified multi-task modeling. By learning generalized failure semantics from unlabeled telemetry, self-supervised representation learning provides a shared diagnostic foundation for AD, FT, and RCL~\cite{sun2024art}. Therefore, we adopt it as our foundational methodology.

\begin{figure*}[t]
\centering
\includegraphics[width=1\textwidth]{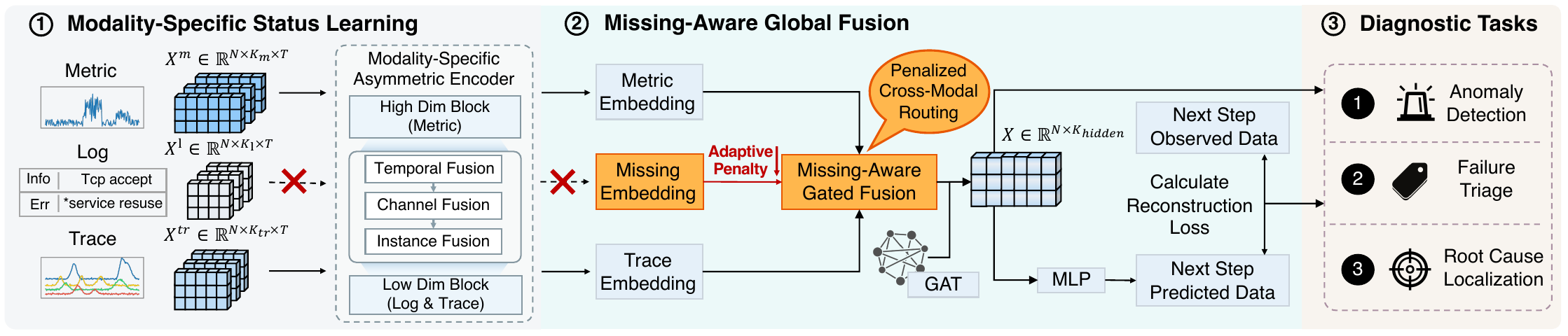}
\caption{Overview of the proposed missing-aware incident management framework. It comprises three core modules: (1) Modality-Specific Status Learning, extracting intra-modality features via an asymmetric encoder; (2) Missing-Aware Global Fusion, integrating isolated embeddings through an attention-guided gating mechanism and a topology-aware graph network; and (3) Diagnostic Tasks, optimizing the network via offline self-supervised reconstruction and executing online AD, FT, and RCL using the unified representations.}
\label{fig:method}
\end{figure*}

\subsection{Missing Modalities in Multimodal Fusion}

While self-supervised unified frameworks have advanced microservice incident management, they universally assume complete modality availability~\cite{sun2024art, sun2025trioxpert, nie2025dest}, leaving them ill-equipped for missing modality scenarios common in production deployments. The multimodal learning community offers two categories of solutions~\cite{wu2026deep}. Generative imputation~\cite{li2025multimodal, li2025simmlm} synthesizes missing channels but incurs substantial computational overhead and risks fabricating false diagnostic signals. Representation-level methods~\cite{xu2024leveraging, mohapatra2025maestro} adjust modality contributions via dynamic routing yet often depend on a primary modality for alignment, failing when key signals drop.

Applying these solutions to microservice observability is complicated by a structural asymmetry absent from standard multimodal tasks. Unlike domains with symmetric latent spaces~\cite{wu2026deep}, microservice data comprises dense, continuous metrics and sparse, discrete logs and traces, whose heterogeneous nature makes uniform imputation strategies unreliable. When metrics are missing, static placeholders such as zero-filling are misread by neural networks as pseudo-normal signals, causing distribution shifts that suppress anomaly indicators in the surviving modalities~\cite{xu2024leveraging}. A reliable framework must therefore isolate modality-specific semantics prior to fusion to prevent imputation noise from corrupting the diagnostic representation.

\subsection{Problem Formulation}

Let $\mathcal{I} = \{1, \dots, N\}$ denote the set of $N$ interconnected microservice instances. For a specific time window, the complete multimodal observation for instance $i$ is defined as $X_i = (X^{\mathcal{M}}_i, X^{\mathcal{L}}_i, X^{\mathcal{T}}_i)$, where $X^{\mathcal{M}}_i \in \mathbb{R}^{T \times d_m}$ represents continuous metric time series, $X^{\mathcal{L}}_i$ denotes semi-structured log sequences, and $X^{\mathcal{T}}_i$ captures distributed trace graphs. The system-wide input is aggregated as $X = \{X_i\}_{i=1}^N$.

To formalize missing modalities caused by infrastructure vulnerabilities, we define an observation mask vector $O = [o^{\mathcal{M}}, o^{\mathcal{L}}, o^{\mathcal{T}}] \in \{0, 1\}^3$, where $o^k=1$ indicates successful collection of modality $k \in \{\mathcal{M}, \mathcal{L}, \mathcal{T}\}$, and $o^k=0$ denotes its absence. The actual incomplete input is $\tilde{X} = \{ \tilde{X}_i \}_{i=1}^N$, with each modality defined as $\tilde{X}_i^k = X_i^k$ if $o^k = 1$, and $\tilde{X}_i^k = \mathbf{0}$ (a structure-compatible static placeholder) otherwise.

Unlike existing methods~\cite{sun2024art, sun2025trioxpert, nie2025dest} assuming $O=[1,1,1]$, we aim to learn a robust mapping $F_\theta: (\tilde{X}, O) \to (y, s, P)$ that explicitly leverages $O$ to mitigate imputation noise. This function jointly optimizes three tasks: (1) AD, predicting a binary indicator $y \in \{0, 1\}$; (2) FT, classifying the failure type $s \in \mathcal{S}$ from a predefined taxonomy; and (3) RCL, estimating a probability distribution $P = [p_1, \dots, p_N] \in [0, 1]^N$ over instances. The core challenge is ensuring $F_\theta$ remains invariant to distributional shifts induced by static placeholders ($\mathbf{0}$) when $o^k=0$, preserving the diagnostic fidelity of available modalities.

\section{Methodology}

\subsection{Overview}
As shown in Figure~\ref{fig:method}, ARMOR operates in two phases. During the \textit{offline} phase, Module 1 first partitions the multimodal input and extracts modality-specific features independently for each modality from anomaly-free operational data, preserving the distinct semantics of continuous metrics and sparse events. Module 2 then integrates these isolated embeddings via a masked gating mechanism and a topology-aware graph network, establishing a normal operating baseline through self-supervised reconstruction so that subsequent anomalous deviations become identifiable. During the \textit{online} phase, the trained model encodes incoming data to obtain latent embeddings alongside next-step predictions; the continuous deviations between predictions and observations are concatenated with the latent embeddings to form unified failure representations. These representations serve as the shared foundation for Module 3, which sequentially monitors for anomalies, determines the failure type upon detection, and localizes the root cause instance.

\subsection{Modality-Specific Status Learning}

\paragraph{Multimodal Data Serialization.} Following established practices~\cite{lee2023eadro, sun2024art}, we transform metrics, logs, and traces into aligned time series. While metrics only require standard normalization, logs are parsed into event frequency series, and traces are aggregated into minute-level statistics (\eg, average latency, request counts). We concatenate these into a fused multivariate time series $M = [M_{metric} \parallel M_{log} \parallel M_{trace}]$ and apply nearest-neighbor interpolation to standardize the resolution to one minute. Finally, z-score standardization~\cite{al2006data} over a sliding window produces the normalized input sequence $X = \{X^{(1)}, \dots, X^{(T)}\}$, where $T$ denotes the window length, and each snapshot $X^{(t)} \in \mathbb{R}^{N \times K}$ encompasses $N$ instances and $K$ channels. This preliminary fusion effectively aligns heterogeneous data~\cite{huang2023twin, zhao2023robust, lee2023eadro, sun2024art, sun2025interpretable}, leaving complex dependency modeling to the subsequent network.

\paragraph{Modality-Specific Feature Extraction.} Directly processing the concatenated $K$ channels overlooks the fundamental \textbf{structural asymmetry} of cloud-native telemetry: metrics reflect macroscopic, continuous resource states, whereas logs and traces record microscopic, discrete execution events. To address this, we partition the sequence into modality-specific sub-matrices $X_{\mathcal{M}}$, $X_{\mathcal{L}}$, and $X_{\mathcal{T}}$. Let $X_m \in \mathbb{R}^{N \times K_m \times T}$ denote the input for modality $m$. We employ a \textit{domain-guided asymmetric encoder} based on ModernTCN~\cite{donghao2024moderntcn} to hierarchically disentangle dependencies. Large-kernel depthwise convolutions first capture long-term temporal trends (e.g., memory leaks):
\begin{equation}
    H_{m, temp} = \text{DepthwiseConv}(X_m).
\end{equation}
Pointwise convolutions then integrate intra-instance channel correlations (e.g., simultaneous latency and I/O spikes):
\begin{equation}
    H_{m, chan} = \text{PointwiseConv}_{chan}(H_{m, temp}).
\end{equation}
Finally, to facilitate early spatial interactions, we permute dimensions and apply instance-level convolution:
\begin{equation}
    H_m = \text{PointwiseConv}_{inst}\Big(\text{Permute}(H_{m, chan})\Big).
\end{equation}
Crucially, applying symmetric extractors to such heterogeneous inputs risks overfitting on sparse events or underfitting on volatile resources. Therefore, our asymmetric design configures high-capacity ModernTCN blocks for continuous metrics to capture intricate fluctuations, while employing lightweight blocks for sparse logs and traces. This physical alignment prevents infrastructure-level noise from overwhelming discrete event semantics. Consequently, the resulting embeddings retain distinct hidden dimensions: $H_{\mathcal{M}} \in \mathbb{R}^{N \times d_{\mathcal{M}}}$, $H_{\mathcal{L}} \in \mathbb{R}^{N \times d_{\mathcal{L}}}$, and $H_{\mathcal{T}} \in \mathbb{R}^{N \times d_{\mathcal{T}}}$, with $d_{\mathcal{M}} > d_{\mathcal{L}}, d_{\mathcal{T}}$.

\subsection{Missing-Aware Global Fusion}

Constructing a comprehensive system representation from isolated unimodal embeddings requires addressing the frequent telemetry outages (\eg, agent crashes or network throttling) prevalent in production environments. Inspired by robust multimodal routing paradigms~\cite{mohapatra2025maestro}, we design a missing-aware global fusion module that employs an attention-guided gating mechanism for incomplete inputs and a graph neural network for spatial service dependencies.

\begin{figure}[t]
    \centering
    \includegraphics[width=1\columnwidth]{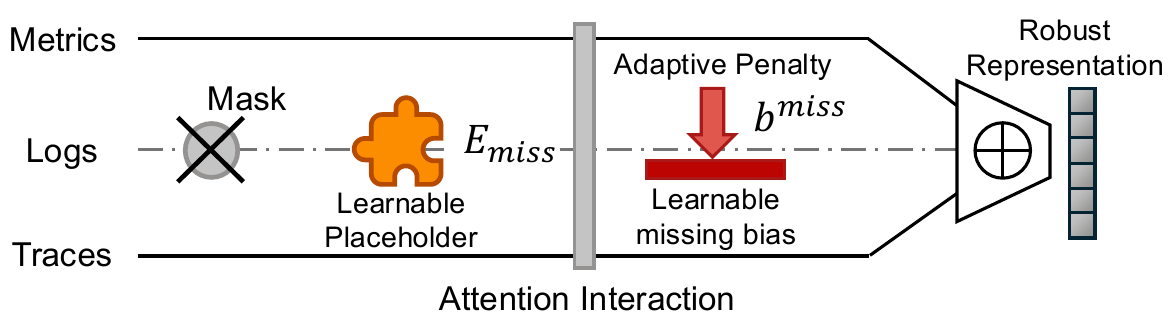}
    \caption{Missing-aware gated fusion. When a telemetry outage occurs (\eg, absent logs), the module first isolates the missing modality by injecting an explicit placeholder ($E_{miss}$). After cross-modal attention interaction, it applies a learnable negative bias ($b^{miss} < 0$) to suppress the routing contribution of the absent modality, ultimately producing a robust instance representation}
    \label{fig:missing_fusion}
\end{figure}

We first project unimodal embeddings $H_{\mathcal{M}}, H_{\mathcal{L}}, H_{\mathcal{T}}$ to a unified dimension $d$, forming a token sequence $V = [v_{\mathcal{M}}, v_{\mathcal{L}}, v_{\mathcal{T}}]$ per instance. A critical challenge arises when a modality collapses: replacing the missing high-dimensional tensor with static zero-padding induces a severe \textit{distribution shift}, causing the network to misinterpret the absence of data as a "healthy idle" state (i.e., pseudo-normal noise). To explicitly isolate this infrastructure-induced artifact, we replace absent tokens ($o_k = 0$) with a \textit{learnable placeholder embedding} $E_{miss} \in \mathbb{R}^d$. Unlike static zeros, $E_{miss}$ acts as a structural \textit{negative signal}, informing the fusion layer of telemetry loss rather than fabricating false stability. We further add a modality-specific embedding $E_{mod, k}$ to preserve semantics. The mask-adjusted token $v'_k$ is formulated as:
\begin{equation}
    v'_k = o_k v_k + (1 - o_k) E_{miss} + E_{mod, k}.
\end{equation}
As shown in Figure~\ref{fig:missing_fusion}, the adjusted sequence $V'$ passes through a multi-head attention (MHA) layer to capture cross-modal interactions:
\begin{equation}
    U = \text{LayerNorm}\Big(V' + \text{Dropout}\big(\text{MHA}(V', V', V')\big)\Big).
\end{equation}

We then dynamically calculate a gating weight for each enriched modality token $U_k$. In a highly dynamic microservice architecture, relying equally on a generalized placeholder and actual telemetry data can derail root cause localization. Since the placeholder lacks the precise execution context of the missing modality, the network must adaptively downgrade its routing contribution and redirect diagnostic attention toward the surviving, reliable signals from other modalities. To enforce this adaptive penalization during cross-modal routing, we compute the gating logit $s_k$ by applying a shared linear projection $W_s \in \mathbb{R}^{1 \times d}$ and adding a learnable missing bias $b_k^{miss} \in \mathbb{R}_{<0}$ (constrained to be negative) exclusively when $o_k = 0$:
\begin{equation}
    s_k = W_s U_k + (1 - o_k) b_k^{miss}.
\end{equation}
The fusion weights $\alpha_k$ are obtained via a softmax function over the logits. Subsequently, the robust instance representation $E \in \mathbb{R}^{N \times d}$ is computed as the weighted sum of the enriched tokens across all modalities, where $N$ represents the total number of microservice instances:
\begin{equation}
    E = \sum_{k \in \{\mathcal{M}, \mathcal{L}, \mathcal{T}\}} \alpha_k U_k.
\end{equation}

Microservice systems operate on complex spatial topologies defined by invocation dependencies. To capture these structural interactions, we model the system as a directed graph $\mathcal{G} = (\mathcal{V}, \mathcal{E})$, where $\mathcal{V}$ represents the microservice instances and $\mathcal{E}$ denotes the interaction links. We apply a multi-layer graph attention network~\cite{2018graph, nie2025dest} to propagate the fused features $E$ along this behavioral graph. Let $\mathcal{N}(i)$ denote the set of topological neighbors for instance $i$, including the instance itself. The graph convolution aggregates spatial features by computing attention coefficients $\alpha_{ij}^{c}$ that quantify the influence of neighbor $j$ on instance $i$. To process information from multiple representation subspaces, we configure the network with a multi-head attention mechanism. The feature update for instance $i$ at the $(l+1)$-th layer is formulated as:
\begin{equation}
    Z_i^{l+1} = \text{ELU} \left( \frac{1}{C} \sum_{c=1}^C \sum_{j \in \mathcal{N}(i)} \alpha_{ij}^{c} W^c Z_j^{l} \right),
\end{equation}
where $Z_i^{l}$ represents the latent feature of instance $i$ at the $l$-th layer, and the initial node feature is set to the fused representation $Z^{0} = E$. The parameter $C$ specifies the number of independent attention heads, while $W^c$ denotes the linear projection matrix for head $c$. To prevent overfitting during spatial propagation, dropout operations are applied to both the feature matrices and the attention coefficients. Finally, the exponential linear unit~\cite{clevert2015fast} activation function is applied after averaging the aggregated outputs from all attention heads.

\subsection{Offline Model Optimization}

Although Figure~\ref{fig:method} primarily illustrates the online diagnostic pipeline, the effectiveness of the unified representations depends entirely on the offline training phase. During this stage, we optimize the entire framework via a self-supervised auto-regression task. To simulate infrastructure-level outages, we apply stochastic modality dropout augmentation: during each training iteration, the binary observation mask $O$ is generated by independently dropping entire modality channels, forcing the framework to infer missing operational states from the surviving signals of topologically correlated microservices. A multi-layer perceptron projects the topology-aware representation $Z^{(T)}$ to predict the full-channel data of the next time step: $\hat{X}^{(T+1)} = \text{MLP}(Z^{(T)})$. By reconstructing both observed and masked modalities, the model learns cross-modal invariants (\eg, inferring CPU trends from trace volumes). The training objective comprises two masked mean squared error terms:
\begin{equation}
    \mathcal{L}_{obs} = \frac{1}{\sum O} \left\| (\hat{X}^{(T+1)} - X^{(T+1)}) \odot O \right\|_F^2,
\end{equation}
\begin{equation}
    \mathcal{L}_{miss} = \frac{1}{\sum (1-O)} \left\| (\hat{X}^{(T+1)} - X^{(T+1)}) \odot (1-O) \right\|_F^2.
\end{equation}
Note that $\mathcal{L}_{miss}$ is computable during training because the offline phase uses complete historical archives; missing channels are produced by stochastic dropout augmentation, not structural absence. The total objective is $\mathcal{L} = \mathcal{L}_{obs} + \lambda \mathcal{L}_{miss}$, with $\lambda$ serving as a balancing weight. For downstream diagnostic tasks, we construct a unified failure signature by concatenating reconstruction errors and latent embeddings as $R^{(t)} = [ (\hat{X}^{(t)} - X^{(t)}) \parallel Z^{(t)} ]$, where $\parallel$ denotes concatenation.

\subsection{Online Diagnostic Tasks}

During online execution, unified representations learned by the self-supervised backbone drive three sequential diagnostic tasks. AD and RCL operate without fault labels, and FT uses failure-type annotations only for a lightweight downstream classifier.

\textbf{Anomaly Detection.} Explicit reconstruction errors often miss subtle deviations in latent topological states. Therefore, we project the unified representation $R^{(t)}$ at each time step $t$ into a scalar system deviation score $S^{(t)}$ to improve sensitivity. Instead of relying on predefined distribution assumptions, we dynamically calculate the detection threshold using the streaming peaks-over-threshold method based on extreme value theory~\cite{siffer2017anomaly}, fitted to normal operating scores~\cite{ma2021jump, ma2018robust, zhao2023robust}. An anomaly is flagged whenever $S^{(t)}$ exceeds this threshold. To distinguish actual system failures from benign fluctuations (\eg, workload spikes), the system generates a formal alert ($y=1$) and routes it to the triage module only if the anomalous state persists across a delay window $W_d$.

\textbf{Failure Triage.} By keeping the upstream encoder frozen, SREs can retrain the failure classifier for new fault categories without retraining the entire model. Prior studies~\cite{sun2024art} typically employ naive mean pooling, which smooths out critical transient spikes. To assess the failure category $s \in \{1, \dots, K\}$, we apply the reconstruction error component $R^{(t)}_{err} = \hat{X}^{(t)} - X^{(t)}$ and construct a system-level representation via statistical pooling: $R_{sys} = \text{Concat}(\text{Mean}(R_{err}), \text{Max}(R_{err}), \text{Std}(R_{err}))$, capturing baseline deviations, peak severities, and volatility. We feed $R_{sys}$ into a lightweight XGBoost~\cite{chen2016xgboost} classifier with sample weighting to handle class imbalance.

\textbf{Root Cause Localization.} Traditional techniques ignoring latent topological semantics are insufficient. Our approach uses the full unified representation $R^{(t)}$ and compares instance-level signatures against the time-aggregated global failure vector $\bar{R}_{sys} = \text{Mean}(\{R^{(t)}\}_{t=1}^{W_f})$ via cosine similarity across the failure window of length $W_f$:
\begin{equation}
    p_i = \frac{1}{W_f} \sum_{t=1}^{W_f} \frac{R_{i}^{(t)} \cdot \bar{R}_{sys}}{\|R_{i}^{(t)}\| \|\bar{R}_{sys}\|}.
\end{equation}
This computation yields a ranked probability distribution $P = [p_1, \dots, p_N]$ across all instances, where higher similarity indicates a greater probability of being the root cause.

\section{Evaluation}
In this section, we address the following research questions:
\begin{enumerate}[leftmargin=1.5em]
    \item \textbf{RQ1:} How well does ARMOR perform in AD, FT, and RCL? 
    \item \textbf{RQ2:} How robust is ARMOR under missing modalities?
    \item \textbf{RQ3:} Does each core component contribute to ARMOR?
    \item \textbf{RQ4:} How do the major hyperparameters influence the performance of ARMOR?
\end{enumerate}

\subsection{Experimental Setup}

\begin{table*}[t]
\centering
\caption{Overall performance comparison of AD, FT, and RCL on datasets $\mathscr{D}_1$ and $\mathscr{D}_2$ under complete data conditions. The best results are highlighted in bold. Hyphens indicate that the method does not support the corresponding task.}
\label{tab:overall_performance}
\resizebox{\textwidth}{!}{
\begin{tabular}{c|c|ccccccccc|ccccccccc}
\toprule
\multirow{3}{*}{\textbf{Type}} & \multirow{3}{*}{\textbf{Method}} & \multicolumn{9}{c|}{\textbf{$\mathscr{D}_1$}} & \multicolumn{9}{c}{\textbf{$\mathscr{D}_2$}} \\
\cmidrule(lr){3-11} \cmidrule(lr){12-20}
& & \multicolumn{3}{c}{\textbf{AD}} & \multicolumn{3}{c}{\textbf{FT}} & \multicolumn{3}{c|}{\textbf{RCL}} & \multicolumn{3}{c}{\textbf{AD}} & \multicolumn{3}{c}{\textbf{FT}} & \multicolumn{3}{c}{\textbf{RCL}} \\
& & Precision & Recall & F1 & Precision & Recall & F1 & Top1 & Top3 & AVG@5 & Precision & Recall & F1 & Precision & Recall & F1 & Top1 & Top3 & AVG@5 \\
\midrule
\multirow{6}{*}{\rotatebox{90}{Multiple}} 
& \textbf{ARMOR} & \textbf{0.925} & \textbf{1.0} & \textbf{0.961} & \textbf{0.946} & \textbf{0.941} & \textbf{0.938} & \textbf{0.821} & \textbf{0.941} & \textbf{0.910} & \textbf{0.993} & \textbf{1.0} & \textbf{0.997} & \textbf{0.882} & \textbf{0.870} & \textbf{0.869} & \textbf{0.815} & \textbf{0.907} & \textbf{0.893} \\
& ART~\cite{sun2024art} & 0.899 & 0.990 & 0.942 & 0.836 & 0.809 & 0.812 & 0.667 & 0.810 & 0.776 & 0.877 & 0.960 & 0.917 & 0.851 & 0.796 & 0.802 & 0.722 & 0.889 & 0.870 \\
& TrioXpert~\cite{sun2025trioxpert} & 0.880 & 0.972 & 0.924 & 0.852 & 0.768 & 0.807 & 0.651 & 0.778 & 0.773 & 0.854 & 0.972 & 0.909 & 0.814 & 0.725 & 0.767 & 0.550 & 0.775 & 0.750 \\
& Eadro~\cite{lee2023eadro} & 0.425 & 0.946 & 0.586 & - & - & - & 0.137 & 0.315 & 0.302 & 0.767 & 0.935 & 0.842 & - & - & - & 0.157 & 0.315 & 0.310 \\
& DiagFusion~\cite{zhang2023robust} & - & - & - & 0.675 & 0.500 & 0.568 & 0.310 & 0.452 & 0.467 & - & - & - & 0.797 & 0.527 & 0.593 & 0.582 & 0.709 & 0.695 \\
& Dejavu~\cite{li2022actionable} & - & - & - & 0.369 & 0.621 & 0.415 & 0.411 & 0.679 & 0.625 & - & - & - & 0.718 & 0.340 & 0.417 & 0.402 & 0.667 & 0.619 \\

\midrule
\multirow{4}{*}{\rotatebox{90}{Single}} 
& Hades~\cite{lee2023heterogeneous} & 0.866 & 0.863 & 0.865 & - & - & - & - & - & - & 0.867 & 0.868 & 0.868 & - & - & - & - & - & - \\
& ChronoSage~\cite{zhang2025integrating} & 0.904 & 0.842 & 0.872 & - & - & - & - & - & - & 0.945 & \textbf{1.0} & 0.972 & - & - & - & - & - & - \\
& MicroCBR~\cite{liu2022microcbr} & - & - & - & 0.667 & 0.796 & 0.717 & - & - & - & - & - & - & 0.629 & 0.678 & 0.636 & - & - & - \\
& DeepHunt~\cite{sun2025interpretable} & - & - & - & - & - & - & 0.796 & 0.905 & 0.897 & - & - & - & - & - & - & 0.731 & 0.893 & 0.873 \\
\bottomrule
\end{tabular}
}
\end{table*}

\subsubsection{Datasets} 
To maintain fairness with existing methods and ensure comprehensive benchmarking, we evaluate ARMOR on two industry-standard datasets ($\mathscr{D}_1$ and $\mathscr{D}_2$) representing distinct microservice architectures. Both datasets encompass continuous metrics, discrete logs, and distributed traces, with each incident containing ground-truth annotations for the timestamp, failure type, and root cause instance.

\textbf{$\mathscr{D}_1$} originates from a cloud-deployed e-commerce simulation replaying authentic historical incidents. It features a 46-node architecture (40 microservices and 6 virtual machines) with 5 distinct system-level failure categories (\eg, CPU, memory, and network faults). The dataset comprises 3,714 normal periods and 210 failure cases. Its extensive multimodal records include 44,858,388 traces, 66,648,685 logs, and 20,917,746 metrics~\cite{sun2024art}.

\textbf{$\mathscr{D}_2$} originates from the International AIOps Challenge 2021\footnote{https://aiops-challenge.com}, representing the core management system of a top-tier commercial bank. It is composed of 18 heterogeneous instances spanning six months of operation. The dataset includes 12,297 normal periods and 133 failure cases across 6 distinct failure types (\eg, JVM and node-level anomalies). Its massive telemetry records comprise 214,337,882 traces, 21,356,870 logs, and 12,871,809 metrics.

Following established chronological evaluation practices~\cite{meng2019loganomaly, sun2024art, huang2023twin}, we split each dataset using the first timestamp in the last 40\% of failure cases as the cutoff point. This ensures temporal continuity between the training (first 60\%) and testing (remaining 40\%) phases.

\subsubsection{Baseline Approaches} 
We compare ARMOR against nine state-of-the-art baselines, which include multi-task frameworks and single-task methods. The multi-task frameworks consist of ART~\cite{sun2024art}, TrioXpert~\cite{sun2025trioxpert}, Eadro~\cite{lee2023eadro}, DiagFusion~\cite{zhang2023robust}, and Dejavu~\cite{li2022actionable}. For single-task methods, we select Hades~\cite{lee2023heterogeneous} and ChronoSage~\cite{zhang2025integrating} for AD, MicroCBR~\cite{liu2022microcbr} for FT, and DeepHunt~\cite{sun2025interpretable} for RCL. The baselines are configured according to the original papers, with dataset-specific adjustments (\eg, window length). To ensure an independent evaluation, we assume known incident timestamps when assessing FT and RCL for the methods that lack AD modules~\cite{sun2024art}.

\subsubsection{Missingness Simulation}
\label{exp:missing_simulation}

Failures in telemetry pipelines can make data unavailable without changing the underlying incident. Since both datasets provide complete multimodal incident archives, our masking protocol keeps the incident, timestamp, failure type, root cause, and telemetry distribution unchanged while varying only telemetry availability.

Complete modality collapse serves as the primary protocol, removing an entire telemetry stream to model severe failures such as agent crashes, network blocks, or observability throttling. During training, stochastic modality dropout randomly drops entire modalities. During evaluation, a binary vector $O \in \{0, 1\}^3$ masks specific modalities, where $O=[0,1,1]$ removes all metric channels. Finer grained protocols characterize degradation below the stream level through random modality availability, channel-level missingness, and element-wise missingness.
 
\subsubsection{Evaluation Metrics} 

We formulate AD and FT as classification tasks. AD is a binary classification problem that identifies whether a failure occurred, whereas FT is a multi-class classification problem that determines the specific failure type. We calculate the precision, recall, and F1-score based on true positive, false positive, and false negative samples. Specifically, these metrics are formulated as $\text{Precision} = TP / (TP + FP)$, $\text{Recall} = TP / (TP + FN)$, and $\text{F1} = 2 \cdot \text{Precision} \cdot \text{Recall} / (\text{Precision} + \text{Recall})$. To handle the imbalanced distribution of the failure types in FT, we report the weighted average F1-score. For RCL, we evaluate the ranking accuracy of the culprit instances by using the $\text{Top@}K$ metric, which is defined as $\text{Top@}K = \frac{1}{N} \sum_{i=1}^{N} (gt_i \in P_{i[1:K]})$. Here, $N$ denotes the total number of evaluated failures, $gt_i$ represents the ground-truth root cause for the $i$-th failure case, and $P_{i[1:K]}$ denotes the top-$K$ predicted candidates. We also compute the average score across the top five results, which is formulated as $\text{Avg@5} = \frac{1}{5} \sum_{K=1}^{5} \text{Top@}K$.

\subsubsection{Implementation} 
 
We implement ARMOR and all the baseline methods by using Python 3.9.13, PyTorch 1.12.1 (CUDA 11.6), and DGL 0.9.0 (CUDA 11.6). All the experiments are conducted on a dedicated server equipped with a 20-vCPU Intel Xeon Platinum 8470Q processor and a single NVIDIA RTX 4090 GPU (24\,GB).

\subsection{RQ1: Overall Performance}

As Table~\ref{tab:overall_performance} shows, ARMOR consistently achieves the highest performance across all tasks on both datasets. Single-task methods (\eg, Hades, ChronoSage, MicroCBR, DeepHunt) are competitive on individual tasks but ignore shared contextual semantics, exacerbating cascading errors. Supervised multi-task frameworks (\eg, Eadro, DiagFusion, DejaVu) rely heavily on historical fault labels, limiting generalization. Self-supervised frameworks (\eg, ART, TrioXpert) reduce label dependency but employ rigid fusion paradigms that fail to accommodate the structural asymmetry between dense continuous metrics and sparse discrete events. In contrast, ARMOR's domain-guided asymmetric encoder and self-supervised reconstruction are aligned with the physical reality of microservice telemetry, effectively mitigating negative transfer across tasks. Across all methods and both datasets, ARMOR achieves the best results on every metric. Particularly in the Failure Triage (FT) task, ARMOR improves over the overall best-performing baseline by 15.5\% in F1-score on $\mathscr{D}_1$ and 8.4\% on $\mathscr{D}_2$. Furthermore, in Root Cause Localization (RCL), it surpasses the strongest multi-task baseline by up to 17.3\% in Avg@5 on $\mathscr{D}_1$.

Beyond diagnostic accuracy, Table~\ref{tab:inference_time} shows that ARMOR consistently achieves lower latency than ART across all tasks on both datasets, confirming that the shared encoder and parallelized fusion introduce minimal overhead for online deployment.

\begin{figure*}[t]
    \centering
    \subfloat[Performance comparison under missing scenarios.]{
        \includegraphics[width=0.75\textwidth]{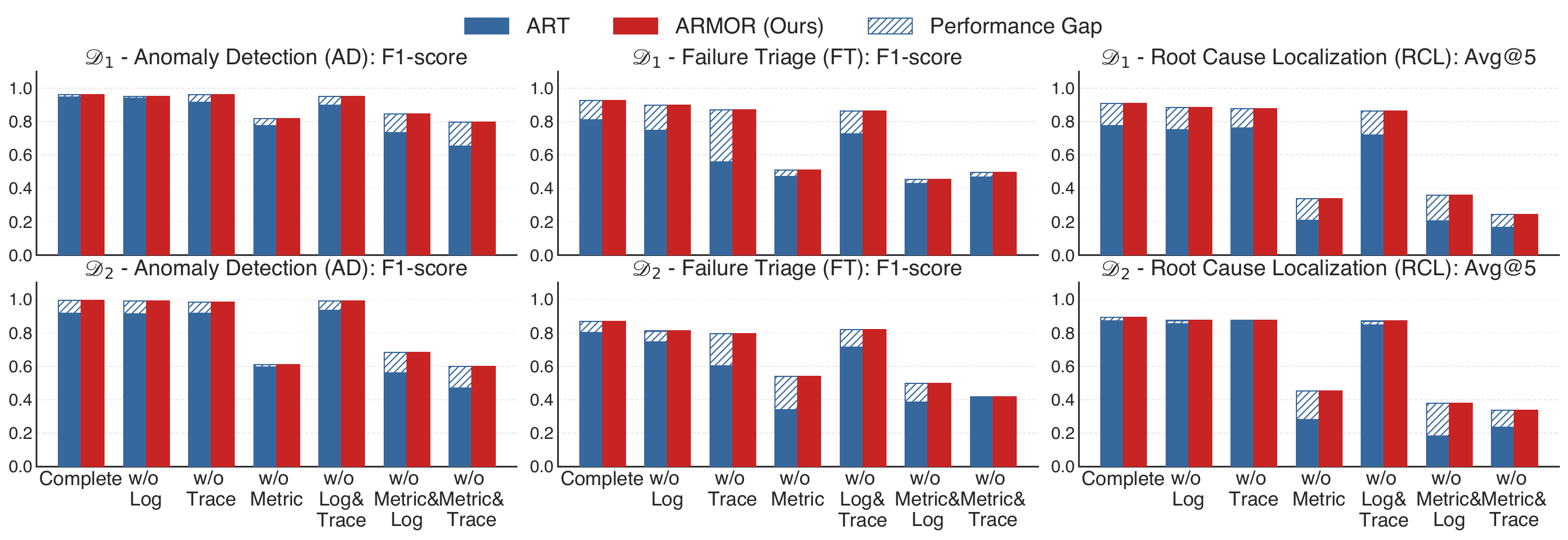}
        \label{fig:robustness_a}
    }
    \hfill
    \subfloat[Learned dynamic bias ($b^{miss}$).]{
        \raisebox{0.2cm}{\includegraphics[width=0.23\textwidth]{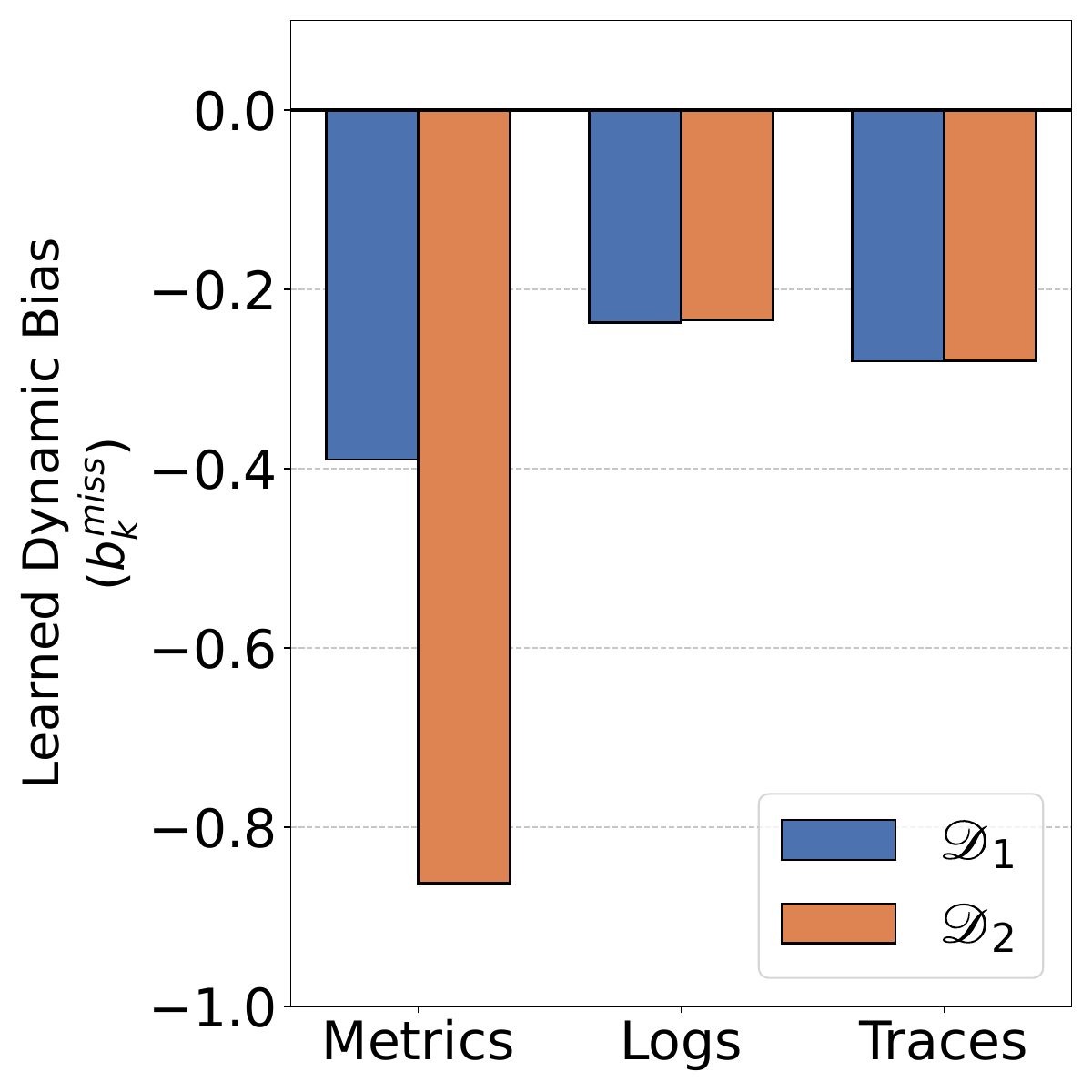}}
        \label{fig:robustness_b}
    }
    \caption{Robustness analysis against missing modalities. (a) Performance comparison across AD, FT, and RCL under various missing modality scenarios on datasets $\mathscr{D}_1$ and $\mathscr{D}_2$. The solid bars represent the diagnostic accuracy of ARMOR and ART, while the hatched regions denote the explicit performance gap indicating the margin by which ART falls behind ARMOR. (b) Visualization of the learned dynamic bias ($b^{miss}$) for missing modalities. The model actively learns negative biases to suppress the routing contribution of absent data, with the penalty being most severe for the high-dimensional metrics.}
    \label{fig:robustness}
\end{figure*}

\begin{figure*}[t]
    \centering
    \includegraphics[width=0.9\textwidth]{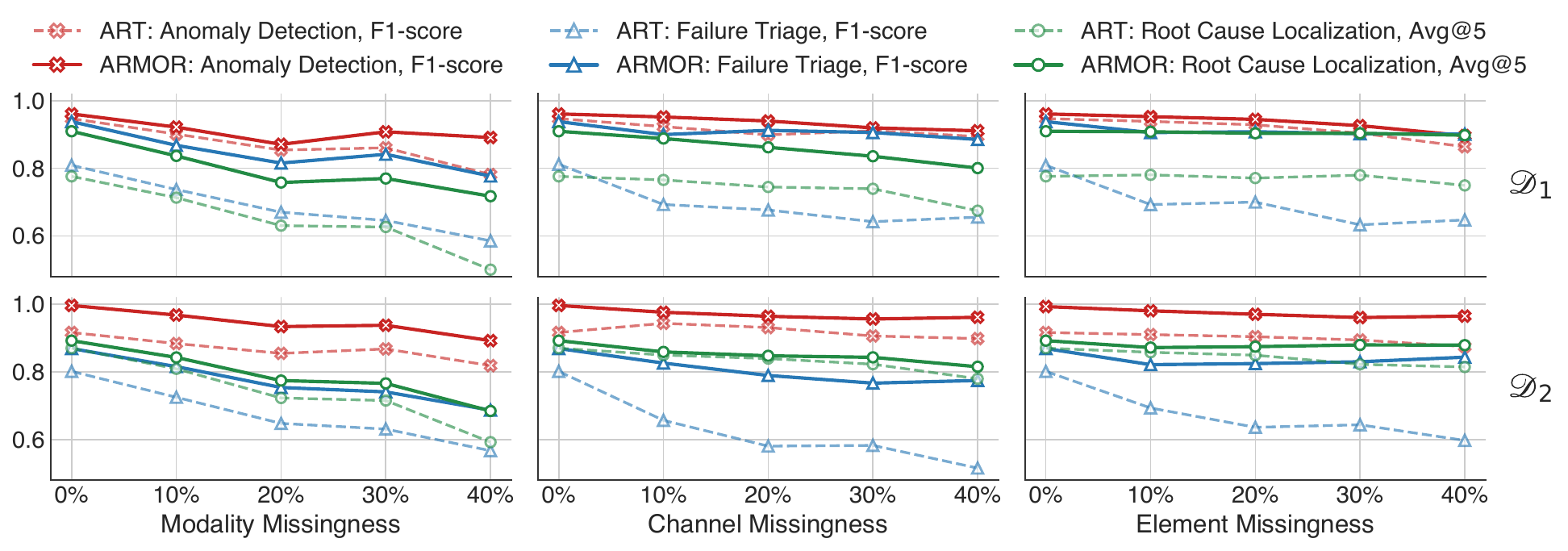}
    \caption{Fine-grained robustness analysis under intermittent modality loss, partial channel degradation, and element-wise telemetry corruption. ARMOR maintains more stable AD, FT, and RCL performance than ART as missingness becomes increasingly severe.}
    \label{fig:fine_grained_missingness}
\end{figure*} 

\subsection{RQ2: Robustness under Missing Modalities}

\subsubsection{Complete Modality Collapse}
\label{sec:complete_modality_collapse}

To evaluate framework robustness against incomplete data, we simulate vulnerabilities in the observability infrastructure by systematically masking specific modality combinations during the online inference phase. We design six distinct missing scenarios that encompass single-modality and dual-modality absences. We focus on ART as the robustness reference: it shares ARMOR's self-supervised orientation and unified task scope, making it the most meaningful counterpart. Supervised baselines require fault labels incompatible with our protocol; single-task methods cover only pipeline subsets. Figure~\ref{fig:robustness_a} compares ARMOR against ART. To ensure a fair comparison, we retrain ART under the same modality dropout augmentation protocol applied to ARMOR, additionally equipping it with three standard imputation strategies (zero-padding, mean imputation, and KNN imputation), and report its best result per scenario. Hatched regions denote the performance gap. ARMOR outperforms ART across all scenarios on both datasets.

The performance degradation exhibits a distinct asymmetry depending on the missing modalities. While the absence of sparse events (such as logs or traces) causes relatively constrained declines, the absence of continuous metrics, which encode critical system states, triggers a severe systemic collapse in the baseline model. Across all varying degrees of data corruption, ARMOR consistently maintains a highly stable functional baseline. The performance gap between the two models widens substantially as more modalities are removed: when the metric modality is absent, ARMOR retains over 60\% higher RCL accuracy than ART across both datasets. This empirical superiority directly validates the proposed missing-aware global fusion mechanism. Because ART is rigidly designed under the assumption of complete data, encountering missing modalities forces the use of naive static padding to maintain input dimensions. This passive imputation introduces pseudo-normal noise that distorts the surviving healthy signals. In contrast, ARMOR inherently accommodates missing data by using learnable placeholders and dynamic gating biases to adaptively redistribute routing attention and isolate the semantic impact of missing data.

Figure~\ref{fig:robustness_b} visualizes the learned dynamic bias $b_k^{miss}$ for each missing modality. Two observations validate our design: (1) all learned biases are strictly negative, confirming that the network actively penalizes placeholder tokens to suppress their routing contributions; (2) the penalty magnitude aligns with structural asymmetry: the model learns a significantly larger negative bias for missing metrics than for sparser logs and traces, as metric placeholders introduce the most severe pseudo-normal noise. This confirms that ARMOR automatically quantifies and mitigates the varying degrees of semantic threat posed by different modality failures.

\subsubsection{Fine Grained Missingness}
\label{sec:fine_grained_missingness}

We further evaluate whether ARMOR remains robust beyond complete modality collapse by introducing three finer grained missingness patterns. Modality missingness randomly removes modalities at different missingness rates while keeping at least one modality available, reflecting intermittent collector or exporter failures. Channel missingness masks feature channels within available modalities, reflecting partial telemetry degradation inside an observability stream. Element missingness randomly removes individual telemetry values, reflecting silent corruption and sporadic data loss. Each setting is repeated with 10 random masks, and Figure~\ref{fig:fine_grained_missingness} reports the average performance as the missingness rate increases from 0\% to 40\%.

Figure~\ref{fig:fine_grained_missingness} shows that ARMOR consistently maintains stronger diagnostic performance than ART across both datasets, all three tasks, and all three missingness granularities. The advantage is especially clear for FT and RCL, where corrupted telemetry directly perturbs failure semantics and instance-level evidence. Across increasing missingness rates, ART exhibits visible degradation under modality, channel, and element missingness, while ARMOR preserves more stable task performance. These results indicate that the missing-aware fusion mechanism generalizes from full-stream outages to intermittent, partial, and fine-grained telemetry loss.

\begin{table}[t]
\caption{Total execution time (s) of the diagnostic pipeline (AD, FT, and RCL) over the entire test set on $\mathscr{D}_1$ and $\mathscr{D}_2$.}
\label{tab:inference_time}
\scriptsize
\centering
\begin{tabular}{c|ccc|ccc}
\toprule
\multirow{2}{*}{\textbf{Method}} & \multicolumn{3}{c|}{\textbf{$\mathscr{D}_1$}} & \multicolumn{3}{c}{\textbf{$\mathscr{D}_2$}} \\
\cmidrule(lr){2-4} \cmidrule(lr){5-7}
& \textbf{AD} & \textbf{FT} & \textbf{RCL} & \textbf{AD} & \textbf{FT} & \textbf{RCL} \\
\midrule
ART~\cite{sun2024art} & 6.81 & 5.05 & 22.46 & 28.81 & 4.01 & 12.91 \\
\textbf{ARMOR} & \textbf{5.23} & \textbf{1.56} & \textbf{21.52} & \textbf{6.71} & \textbf{1.45} & \textbf{8.08} \\
\bottomrule
\end{tabular}
\end{table}

\subsection{RQ3: Ablation Study}

To evaluate the key technical contributions of ARMOR, we design twelve variants grouped into two categories. The first category assesses the modality-specific status learning and missing-aware global fusion. Specifically, variants A1, A2, and A3 remove the temporal, channel, and instance fusions, respectively. Variant A4 replaces the missing-aware gated fusion with a naive mean fusion. Variant A5 replaces the learnable missing placeholder with static zero-padding. Variant A6 trains the model exclusively on complete modalities, removing missing modality augmentation. Variants A7, A8, and A9 modify the graph network by removing it entirely, replacing it with GCN, or replacing it with GraphSAGE, respectively. The second category evaluates the unified representation strategies for downstream tasks. In this category, B1 and B3 remove the latent embeddings for AD and RCL, respectively, whereas B2 removes the standard deviation feature for FT.

\begin{table}[t]
\centering
\caption{Ablation study on datasets $\mathscr{D}_1$ and $\mathscr{D}_2$. Variants A1-A9 evaluate the encoder and global fusion components. Variants B1-B3 evaluate task-specific unified representation strategies. Hyphens indicate that the variant evaluates a specific downstream task and remains identical to the base model for the other tasks. The best results are highlighted in bold.}
\label{tab:ablation_study}
\resizebox{\linewidth}{!}{
\begin{tabular}{c|ccc|ccc}
\toprule
\multirow{2}{*}{\textbf{Variant}} & \multicolumn{3}{c|}{\textbf{$\mathscr{D}_1$}} & \multicolumn{3}{c}{\textbf{$\mathscr{D}_2$}} \\
\cmidrule(lr){2-4} \cmidrule(lr){5-7}
& \textbf{AD (F1)} & \textbf{FT (F1)} & \textbf{RCL (Avg@5)} & \textbf{AD (F1)} & \textbf{FT (F1)} & \textbf{RCL (Avg@5)} \\
\midrule
A1 & 0.956 & 0.897 & 0.902 & 0.983 & 0.843 & 0.874 \\
A2 & 0.937 & 0.915 & 0.883 & 0.980 & 0.787 & 0.885 \\
A3 & 0.937 & 0.914 & 0.891 & 0.987 & 0.861 & 0.870 \\
A4 & 0.942 & 0.926 & 0.900 & 0.990 & 0.794 & 0.878 \\
A5 & 0.939 & 0.887 & 0.875 & 0.964 & 0.771 & 0.856 \\
A6 & 0.946 & 0.883 & 0.860 & 0.941 & 0.822 & 0.885 \\
A7 & 0.960 & 0.895 & 0.857 & 0.973 & 0.831 & 0.867 \\
A8 & 0.946 & 0.895 & 0.829 & 0.959 & 0.859 & 0.882 \\
A9 & 0.933 & 0.927 & 0.860 & 0.987 & 0.865 & 0.889 \\
\midrule
B1 & 0.956 & - & - & 0.964 & - & - \\
B2 & - & 0.908 & - & - & 0.830 & - \\
B3 & - & - & 0.767 & - & - & 0.889 \\
\midrule
\textbf{ARMOR} & \textbf{0.961} & \textbf{0.938} & \textbf{0.910} & \textbf{0.997} & \textbf{0.869} & \textbf{0.893} \\
\bottomrule
\end{tabular}
}
\end{table}

\begin{figure*}[t]
  \centering
  \includegraphics[width=0.85\textwidth]{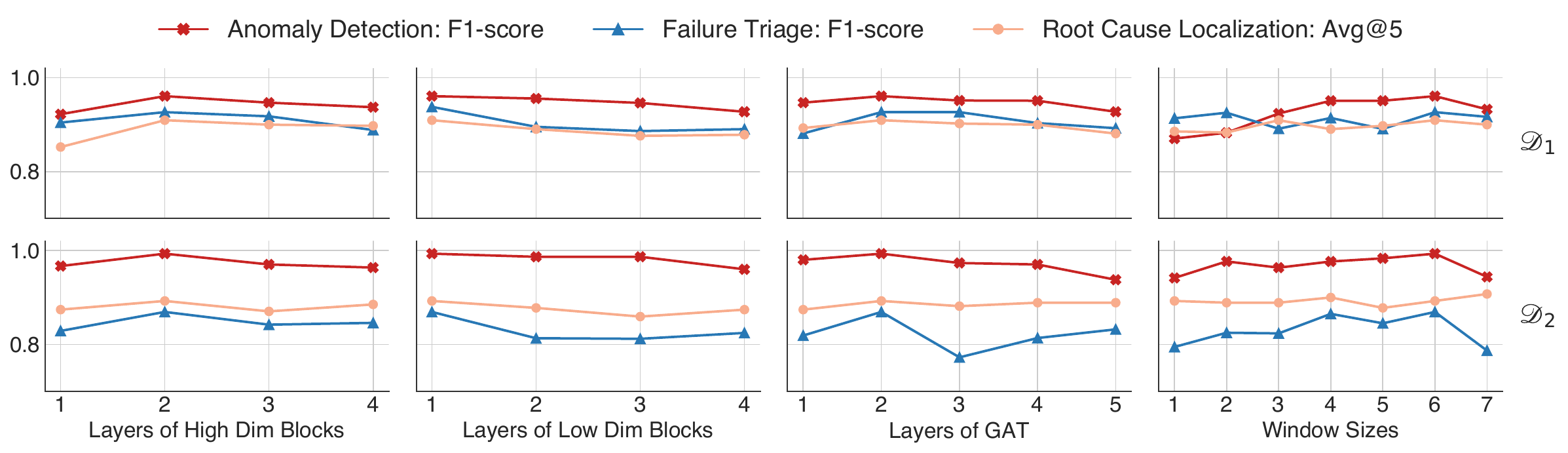}
  \caption{Sensitivity analysis of four core hyperparameters (Layers of High Dim Blocks, Layers of Low Dim Blocks, Layers of GAT, and Window Sizes) across AD, FT, and RCL on datasets $\mathscr{D}_1$ and $\mathscr{D}_2$.}
  \label{fig:hyperparameters}
\end{figure*}

Table~\ref{tab:ablation_study} shows that ARMOR outperforms all variants on both datasets. Performance drops when removing the temporal, channel, or instance fusions (A1--A3), confirming that hierarchical extraction is essential for capturing complex telemetry dynamics. The decline in A4 proves that naive mean fusion fails to suppress imputation noise, validating the gating mechanism. The drops in A5 and A6 confirm our two-pronged missing modality strategy: static zero-padding misleads the network with pseudo-normal signals (A5), and training without modality dropout prevents robust representation learning (A6). Replacing or removing the graph network (A7--A9) degrades localization, indicating that attention-guided spatial propagation better models heterogeneous invocation dependencies than convolutional aggregators. Modifying the task-specific representations (B1--B3) consistently reduces performance, confirming that reconstruction errors, deviation volatility, and topology-encoded embeddings each provide indispensable diagnostic signals for their respective tasks.

\subsection{RQ4: Hyperparameter Sensitivity}
As Figure~\ref{fig:hyperparameters} illustrates, for High Dim Blocks and GAT, performance peaks at moderate depth and degrades beyond it due to overfitting and over-smoothing; Low Dim Blocks are optimal at depth 1, reflecting the simpler structure of sparse modalities. This asymmetry is consistent with the structural difference between dense continuous metrics and sparse discrete logs and traces, where deeper extraction benefits the former but hurts the latter. Window size $T$ follows the same trend: too small misses failure dynamics, too large introduces noise, with $T=6$ providing the best balance. These results confirm that ARMOR is robust to reasonable hyperparameter choices without dataset-specific tuning.

\section{Discussion}

\subsection{Limitations and Possible Solutions}
ARMOR's offline training assumes that historical records are structurally complete enough to serve as reconstruction targets for $\mathcal{L}_{miss}$. In the rare case where a modality has been persistently absent from the archive to an extent beyond the stochastic dropout simulated during training, the missing-aware fusion module loses its explicit training signal. Replacing strict numerical reconstruction with cross-instance contrastive learning, aligning incomplete instances against fully observed peers, offers a viable path forward without requiring perfect historical targets. A second limitation is topology evolution: the dependency graph is built from historical records, so frequent instance creation or destruction requires periodic updates; incremental graph learning is a natural extension.

\subsection{Threats to Validity}
One threat concerns construct validity. Section~\ref{sec:complete_modality_collapse} evaluates complete modality collapse, which mirrors the stream-level blackouts described in Section~\ref{motivation:missing_exist}, where process crashes and network blocks can eliminate an observability stream at once. Section~\ref{sec:fine_grained_missingness} further evaluates modality availability changes, channel-level missingness, and element-wise missingness, showing that ARMOR remains robust across finer granularities of observability degradation. Another construct threat concerns the completeness of historical training archives. Both benchmark datasets provide complete multimodal records during the normal training period, so long-running training with naturally incomplete historical archives remains future work. A second threat concerns internal validity. To rule out imputation bias, we retrain ART under the same modality dropout augmentation protocol as ARMOR, additionally equip it with zero-padding, mean imputation, and KNN imputation, and report its best result per scenario. The observed gap therefore reflects architectural differences rather than training asymmetry. A third threat concerns external validity. Both datasets originate from controlled testbed environments with fewer instances than large-scale industrial deployments and may not capture topology evolution, concept drift, cascading failures, or timing-sensitive incidents. Extending evaluation to live industrial traces remains important future work.

\section{Related Work}

\textit{Single-Task Incident Diagnosis.} Early approaches address individual incident management stages, such as AD \cite{audibert2020usad, chen2024lara, chen2024cluster, huang2023twin, lee2023heterogeneous, zhang2025integrating, kim2025causality, cho2025structured, ding2025enhancing}, FT \cite{ma2020diagnosing, tao2024giving, sun2025failure, sui2023logkg}, and RCL \cite{somashekar2024gamma, lin2024root, pham2024root, han2025root, sun2025interpretable}, in isolation. Although effective locally, isolating these related tasks ignores shared diagnostic contexts, increases maintenance overhead, and renders independent pipelines vulnerable to severe error cascading \cite{sun2024art}.

\textit{Unified Multi-Task Frameworks.} Recent multi-task approaches extract shared knowledge from multimodal data (metrics, logs, and traces) to construct unified frameworks that overcome these limitations. Supervised multi-task models~\cite{lee2023eadro, ding2025adaptive} integrate heterogeneous events but require extensive fault-labeled data throughout training, limiting practical deployment in environments where labeled fault data is prohibitively expensive. To reduce label dependency, self-supervised unified approaches~\cite{sun2024art, sun2025trioxpert, nie2025dest} employ cascaded representation learning to jointly support AD, FT, and RCL. However, existing self-supervised unified frameworks assume perfectly complete observability data, leaving them without a principled missing-aware strategy for production deployment. ARMOR addresses this gap by combining self-supervised representation learning with explicit architectural support for modality absence.

\textit{Missing Modality Learning.} Handling incomplete multimodal inputs has been studied in general settings through generative imputation~\cite{wang2023multi, yao2024drfuse} and representation-level routing~\cite{xu2024leveraging, li2025multimodal, mohapatra2025maestro, li2025simmlm, wu2026deep}. However, these methods either introduce substantial computational overhead from data synthesis or rely on a primary modality for alignment, which fails when the most information-dense stream drops. More critically, microservice observability data exhibits structural asymmetry between dense continuous metrics and sparse discrete logs and traces, a property that general missing modality methods do not account for. To our knowledge, ARMOR is the first self-supervised framework to address missing modalities in the unified incident management setting, treating observability gaps as a first-class design concern rather than a post-hoc fallback that requires fault-labeled training data.

\section{Conclusion}

This paper presents ARMOR, a robust self-supervised framework for automated incident management under missing modalities. ARMOR employs a modality-specific asymmetric encoder and a missing-aware gated fusion mechanism with learnable placeholders and dynamic bias compensation to isolate imputation noise. The unified failure representations concurrently support label-free AD and RCL, with FT requiring only failure-type annotations for a lightweight downstream classifier. Extensive evaluations demonstrate state-of-the-art performance under complete data and strong robustness under severe modality loss. Future work will investigate cross-instance contrastive learning to reduce reliance on complete historical archives.

\section{Data Availability Statement}
The source code and data of ARMOR are available in Zenodo ~\cite{zenodo_armor}.

\section*{Acknowledgments}

This work is supported by the Young Scientists Fund-Type A of the National Natural Science Foundation of China (62125206), the Young Scientists Fund-Type C of the National Natural Science Foundation of China (62502441), and the Major Program of the National Natural Science Foundation of Zhejiang Province (LD25F020002).

\bibliographystyle{ACM-Reference-Format}
\bibliography{ref}

\end{document}